\documentclass[letterpaper, 10pt, twocolumn]{article}
\usepackage{clean}
\shorttitle{Human Sensorimotor Model Adaptation}

\usepackage[caption=false,font=footnotesize]{subfig}
\usepackage{stfloats}

\usepackage{verbatim}

\usepackage{graphicx}
\usepackage{subfig}
\usepackage{cite}

\usepackage{graphics} % for pdf, bitmapped graphics files
\usepackage{epsfig} % for postscript graphics files
\usepackage{bm,times} % assumes new font selection scheme installed
\usepackage{amsmath} % assumes amsmath package installed
\usepackage{amssymb}  % assumes amsmath package installed
\usepackage{amsfonts}
\usepackage{verbatim} %comment
\graphicspath{{figs/}}
\usepackage{epstopdf}
\usepackage{stfloats}

\usepackage{float}
\usepackage{color}
\usepackage{url}
\usepackage{verbatim}

\usepackage[colorlinks=true, linkcolor=black, urlcolor=cyan, filecolor=black, citecolor=black]{hyperref}

\usepackage{booktabs}

\usepackage{float}

\usepackage{url}

\begin{document}

\title{{\LARGE \bf
Human Sensorimotor Model Adaptation }
\\ { \large to Account for Robot Accuracy and Speed Augmentation in Human-Robot Co-Manipulation}
}

\author{Frits van Hall$^1$, Kevin Haninger$^{2}$, and Luka Peternel$^1$
\thanks{$^1$Cognitive Robotics, Delft University of Technology, Delft, The Netherlands (e-mail: fritsvanhall@gmail.com, l.peternel@tudelft.nl)}
\thanks{$^2$Department of Automation, Fraunhofer IPK, Berlin, Germany (e-mail: kevin.haninger@ipk.fraunhofer.de)}
}

\maketitle

\begin{abstract}
In human-robot co-manipulation, the robot's trajectory should accommodate the human's, where human sensorimotor behavior is affected by the task's spatial and temporal constraints, as well as personal factors. Several robot trajectory planning methods have been proposed to embed human sensorimotor models in the robot trajectory planning, but these methods do not consider temporal constraints that adapt to how the human behaves when interacting with the robot. To better align human and robot collaborative motion, this work adapts a stochastic optimal control problem with a final state constraint from the task constraints and a temporal constraint from Fitts' law. A recursive least squares method is used to estimate the Fitts' law parameters, adapting the robot's motion duration to that of the human, thus allowing the robot to adapt to how the human behaves when interacting with the robot. It is hypothesized that a speed increase can result from the robot augmenting the accuracy. The feasibility of the approach is validated on a positioning task where the robot controls degrees of freedom which have repeatable motion, and the human adapts motion in the complementary degrees of freedom. The approach is shown to be computationally suited to online use, it converges as the human and robot mutually adapt, and can increase the movement speed, although thorough human studies are needed in future work.  
\end{abstract}

% \begin{IEEEkeywords}
% 
% \end{IEEEkeywords}

%\IEEEpeerreviewmaketitle

\section{Introduction}

\begin{figure}[t]
    \centering
    \includegraphics[width=0.80\linewidth]{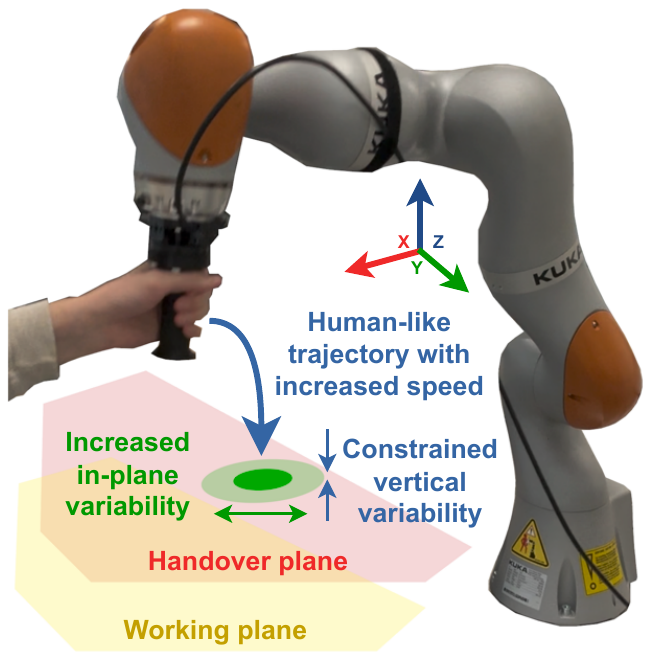}
    \caption{The proposed sensorimotor planner generates a human-like trajectory (blue) to an authority handover plane (red), where the robot controls the vertical z-axis motion with a high impedance, and the human controls the horizontal x/y-plane motion. By including the known handover plane as a final state constraint, the robot eliminates the primary source of task risk, which is collisions with the working plane (e.g., assembly plane) (yellow). The exact position within the plane (dark green) is determined by the human to adapt to various goal positions (diverging arrows, light green). The robot's motion should be synchronized so the human doesn't need to wait for the robot, even when a new goal location is found. The proposed planner includes a Fitts' law-based temporal constraint, adapting it online using observed movement durations and distances.}
    \label{fig:overview}
    %\vspace{-4mm}
\end{figure}

Human presence is still essential in many scenarios involving humanoid robots, where high adaptability is paramount, such as in high-mix/low-volume manufacturing, inspection \& maintenance, and household assistance. Human-robot collaboration is a way to combine complementary strengths of humans and humanoid robots to work toward common goals~\cite{ajoudani2018progress}. Robots can provide precision and load-bearing capacity, while humans contribute perception, task knowledge, and cognitive flexibility~\cite{mol2025fitts}. To achieve such seamless and efficient collaboration, robots must develop an understanding of human intentions, behavior, and preferences. This can be achieved with robot learning from the collaborating humans~\cite{losey2022physical,semeraro2023human,haninger2023model}. Data-driven approaches using machine learning can personalize the learned models to a specific human. However, high-quality data from high-mix low-volume industrial settings are scarce, and learned models lack interpretability, making it hard to adapt them to new tasks without retraining~\cite{hoffman2024inferring}.

An alternative to machine learning is to directly integrate the established human motor control knowledge into robotic systems~\cite{fang2023human}. Unlike data-driven approaches, theory-driven human motor control models are based on observable and testable phenomena, yield interpretable predictions, and are more generalizable. Once identified, the personalization only requires tuning a few parameters. Thus, human motor control models present an underexplored opportunity for human-robot collaboration, where insights about how the central nervous system (CNS) plans the movement of the human limbs in goal-directed tasks can help coordinate robot assistance.

The computational human motor control model typically prescribes a mathematical (objective) function, where the CNS plans movements that optimize various aspects captured by that function. The objective function is then fitted to and tested with observed human movements from the empirical data. In unconstrained conditions, human movements generally tend to exhibit slightly curved hand paths in Cartesian space with smooth, bell-shaped velocity profiles~\cite{flash1985coordination,franklin2011computational,peternel2017unifying}. Moreover, human motor control is subject to signal noise (both in commands and feedback), and is therefore inherently a stochastic process~\cite{harris1998signal}.

One of the earliest computational models is the minimum-jerk model~\cite{flash1985coordination}, which assumes that humans minimize the time integral of squared jerk. While this model captures some of the key aspects observed in human goal-directed movements, e.g., smooth trajectories, it is deterministic and does not directly capture movement speed adjustments for different accuracy requirements. A well-known Fitts' law~\cite{fitts1954information} that aligns with observations of human movements states that faster movements are inevitably less accurate, which results from stochasticity and noise. Computational models that can directly capture Fitts' law are called speed-accuracy trade-off (SAT) models. SAT models account for signal-dependent noise in the arm dynamics, where the noise magnitude scales proportionally with the control signal~\cite{harris1998signal,todorov2002optimal}. Intuitively, the faster we move, the more noise we induce with higher muscle activity, which results in more movement variability and lower accuracy.

While pure SAT models account for Fitts' law, they do not directly incorporate the cost of movements in terms of effort spent and the perceived benefit that is discounted by the time invested in them. The models that incorporate those two factors are called cost-benefit tradeoff (CBT) models~\cite{shadmehr2010temporal,rigoux2012model}. It is also possible to combine both SAT and CBT into a unified computational model~\cite{peternel2017unifying}, which allows for capturing both the spatial and temporal characteristics of human reaching within a single cost function. The unified model prescribes variability of endpoint position at the target and permits a few misses in repeated task execution at the expense of faster execution, since slower movements are perceived as less beneficial. The model proposed in~\cite{berret2021stochastic} penalizes both endpoint inaccuracy and control effort in its objective function. The formulation enforces a final state constraint, ensuring movements end exactly at point targets. This constraint suppresses the final endpoint variability that would otherwise arise from signal-dependent noise. The movement duration is treated as a free variable and emerges from the optimization itself, rather than explicitly from SAT. 

The unified SAT and CBT model has been successfully applied for human intent recognition in human-robot collaboration for co-manipulation tasks~\cite{haninger2024planning}. While this was useful to synchronize human and robot movements based on inherent human capabilities, it did not allow the robot to enhance the human beyond their limitations. Ideally, we would like the robot to enable humans to operate beyond the inherent SAT and move faster without losing accuracy. While the model in~\cite{berret2021stochastic} provides the structure for precision through final state constraints, it lacks explicit consideration of target width. More critically, neither model accounts for how human behavior adapts when the robot itself alters the task's risk profile through accuracy augmentation.

To understand this adaptation, let us consider a common co-manipulation scenario in assembly applications, where a sensorimotor planning model generates robot-assistive movements and synchronises them with the human during goal-directed actions (Fig.~\ref{fig:overview}). In typical manufacturing scenarios, the working plane's position (e.g., assembly surface) is known to the robot in advance, but the exact target position on the working plane remains unknown, so the system exploits human cognitive capabilities to determine it. The robot planner can then generate trajectories toward a known authority handover plane, positioned in front of the working plane, where the human can take over for the final cognitively demanding assembly action. The impedance controller then maintains high stiffness orthogonal to this authority handover plane, where the robot dominates the assistive movement to augment the human speed, and low stiffness within it, where the human determines the exact assembly point relying on vision and task knowledge~\cite{haninger2024planning}. This strategy directly instantiates the HRC vision of leveraging complementary strengths. The robot contributes precision by guaranteeing safety and accuracy in the high-risk direction, while the human contributes perception and task knowledge to retain flexible control within the plane.

This split in responsibilities could enable faster human movement. Since the robot suppresses variability in the main movement direction, the remaining probability of missing the target lies primarily in the plane, where small positional errors carry minimal consequences. These in-plane errors are tolerable because the authority handover plane is positioned before the assembly plane, leaving sufficient distance for a corrective sub-movement to reduce accumulated error and reach the final target on the assembly plane~\cite{elliott2001century,peternel2019target}. The robot's accuracy augmentation effectively increases the acceptable probability of missing the target in the plane dimensions. Due to the SAT, this loosened accuracy requirement enables humans to prioritize speed at the expense of in-plane precision, allowing for faster movement and task completion. This necessitates a planner capable of adapting to this augmentation-enabled speed increase. However, no existing stochastic sensorimotor planner for robotics simultaneously addresses both challenges of suppressing variability through a final state constraint for accuracy augmentation, while adapting to the expected increase in human speed resulting from this augmentation.

To address these gaps, an accuracy- and speed-augmenting planning framework for co-manipulation has been developed. We used the model ~\cite{berret2021stochastic} as a foundation, as it allows a final state constraint to suppress variability for accuracy augmentation (see Appendix for comparison with the model from~\cite{peternel2017unifying}). We exploit the model's independent control over movement duration to extend the model with a temporal adaptation mechanism that can incorporate SAT. This was done by a horizon-constraint based on Fitts' law that predicts movement duration from task difficulty, determined by predicted target distance and width. To estimate how fast the human-robot system completes movements at a set difficulty, we created an adaptive estimation scheme using Recursive Least Squares (RLS). This insight enables adaptive planning for human augmentation by observing actual movement durations and distances. Furthermore, to achieve real-time computation needed for human-robot co-manipulation, the original stochastic optimal control framework from ~\cite{berret2021stochastic} has been scaled and optimized to higher degrees of freedom for robotic use.

The research objectives/contributions of this study are:
\begin{itemize}
\item O1: Develop a sensorimotor planner for co-manipulation that generates human-like reaching trajectories and can synchronize with the human's motions.
\item O2: Adapt movement duration to the task difficulty through a Fitts' law-based temporal constraint.
\item O3: Enable the planner to account for the expected faster human movements that result from robotic accuracy augmentation.
\item O4: Validate the feasibility of the proposed method by showing that it adapts speed to task difficulty according to Fitts' law while maintaining endpoint precision, enabling robotic accuracy augmentation.
\end{itemize}

\begin{figure}[!t]
    \centering
    \includegraphics[width=1\linewidth]{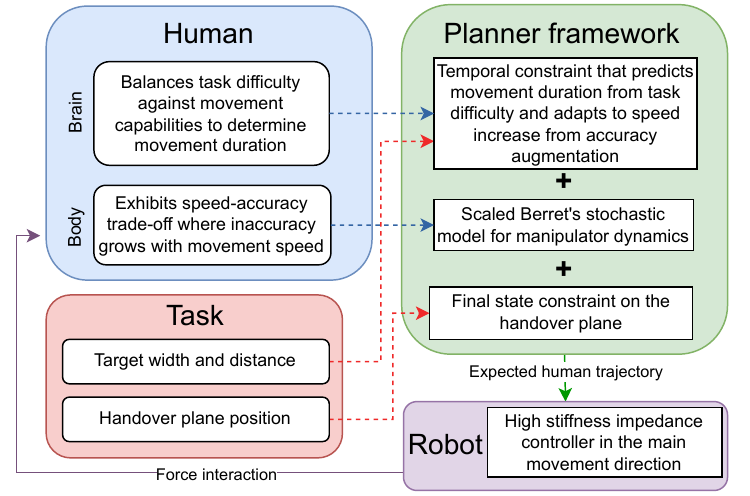}
    \caption{Overview of the proposed sensorimotor planning framework. The planner models aspects of human motor control and task constraints. Dashed lines indicate which aspect the planner components infer or model. The temporal constraint models how humans determine movement duration from task difficulty and adapts to speed increases from robotic augmentation; the scaled stochastic model embodies human motor control principles to generate smooth robot trajectories; the final state constraint exploits the known authority handover plane position to suppress endpoint variability for accuracy augmentation. The solid line represents the physical force interaction between the impedance controller and the human body during co-manipulation.}
    \label{fig:overview_pet_inspired}
    %\vspace{-4mm}
\end{figure}

\section{Methods}
Figure~\ref{fig:overview_pet_inspired} presents a contextual overview of the proposed method. The central component of the method is the human sensorimotor planner framework. Within this framework, the stochastic optimal control formulation predicts human movements according to the defined objective function. The final state constraint ensures that reaching the target (i.e., the hand-over plane) is respected. The temporal constraint enables explicit incorporation of SAT through Fitts' law. Finally, the iterative Fitts' law parameter adaptation mechanism allows online adjustment of human capabilities/limitations when they are augmented by the robot during co-manipulation.

\subsection{Framework and Objective Function}
We used a stochastic modeling framework by treating the system's state at each timestep as a Gaussian random variable $\bm{x}_k \sim \mathcal{N}(\bm{\mu}_k, \bm{\Sigma}_k)$. This linear discrete-time approximation propagates the state mean $\bm{\mu}_{k+1}$ and covariance $\bm{\Sigma}_{k+1}$ forward in time, with signal-dependent noise injected through the covariance propagation to capture how control effort affects endpoint variability.

For the reaching task under consideration, the control objective is to move the robot from an initial state $\bm{x}_0$ with zero initial velocity to reach a target end-effector position $\bm{p}_{target}$ in $N$ time steps while minimizing end-effector variance, control effort, and trajectory deviation from a straight path. The objective function consists of terms related to the terminal cost and the running cost, defined as
\begin{equation}
    J(\bm{u}_{0:N-1})
    = r\,\Phi(\bm{\mu}_N,\bm{\Sigma}_N)
      + \sum_{k=0}^{N-1} l(\bm{\mu}_k,\bm{u}_k),
      \label{eq:total_cost}
\end{equation}
where $r$ is a weight balancing the two terms, $\Phi(\bm{\mu}_N,\bm{\Sigma}_N)$ penalizes the final end-effector variance, and $l(\bm{\mu}_k,\bm{u}_k)$ is the running cost at each timestep. For computational efficiency, we propagate only the diagonal of the state covariance (variances $\boldsymbol{\sigma}^{2}$) rather than the full covariance matrix, thus ignoring correlations between joints.

The terminal cost is
\begin{equation}
  \Phi(\bm{\mu}_N,\bm{\Sigma}_N) =
     \,\operatorname{tr}\!\left(
      \bm{J}(\bm{\mu}_N)\,
      \operatorname{Diag}\!\big(\boldsymbol{\sigma}^{2}_{q,N}\big)\,
      \bm{J}(\bm{\mu}_N)^\top
    \right),
    \label{eq:terminal_cost}
\end{equation}
where $\operatorname{Diag}\!\big(\boldsymbol{\sigma}^{2}_{q,N}\big)$ forms a diagonal matrix from the final joint position variances, and $w_\Sigma$ weights the terminal cost. The trace operation sums the diagonal elements of the resulting end-effector covariance matrix, yielding the total predicted end-effector variance.

The running cost penalizes two terms: torque rate and deviation from a straight-line path in task space, defined as
\begin{equation}
    l(\bm{\mu}_k,\bm{u}_k) = 
    w_u \,\|\dot{\boldsymbol{\tau}}_k\|^2 
    \;+\;
    w_{\perp}\,\| P_\perp(\bm{p}_k - \bm{p}_0)\|^2,
    \label{eq:running_cost}
\end{equation}
where $\dot{\boldsymbol{\tau}}_k = \bm{u}_k - \bm{u}_{k-1}$ is the torque rate of change, which encourages smooth control inputs.

The straight-path term penalizes deviations from the direct line defined by the unit target direction vector $\bm{v} = \bm{p}_{\mathrm{target}} - \bm{p}_0$, where $\bm{p}_0 = \mathrm{FK}(\bm{\mu}_{q,0})$ is the initial end-effector position and $\bm{p}_{\mathrm{target}}$ is the target position. The orthogonal projection operator is $P_\perp = I_3 - \hat{\bm{v}}\,\hat{\bm{v}}^\top$, which projects vectors onto the plane orthogonal to $\hat{\bm{v}}$ and $\hat{\bm{v}} = \frac{\bm{v}}{\|\bm{v}\|}$. At each timestep, the current end-effector position $\bm{p}_k = \mathrm{FK}(\bm{\mu}_{q,k})$ is used in $P_\perp(\bm{p}_k - \bm{p}_0)$, which penalizes the component of the displacement orthogonal to $\hat{\bm{v}}$. This straight-path cost was necessary when scaling the framework to higher-dimensional robotic systems, to prevent curved trajectories that deviated significantly from the direct paths typical of human reaching.

\subsection{Final state constraint}
The objective function in a stochastic framework generates trajectories with natural endpoint variability resulting from signal-dependent noise injection. However, this formulation does not suppress endpoint variability, which is necessary to ensure the trajectory reaches the target precisely. For accuracy-augmenting co-manipulation, we require the robot to end precisely at the authority handover plane to suppress variability in the critical movement direction. To do this, following~\cite{berret2021stochastic}, we impose hard equality constraints on the final mean state. 

Determining the target position for the final mean state constraint presents a challenge, as the exact in-plane target position on the working plane (e.g., where on the assembly surface) is determined by the human during the movement based on visual feedback and task knowledge. As explained in the introduction, we exploit the known authority handover plane position by directing the robot toward this plane positioned in front of the working plane. While the human retains control over the final in-plane position, the planner requires a specific target position for the mean state constraint. The target position $\bm{p}_{target}$ on the authority handover plane is defined as the closest point from the initial end-effector position to the handover plane. The final state constraint at step $N$ is $\mathrm{FK}(\bm{\mu}_{q,N}) = \bm{p}_{target}$, where $\mathrm{FK}(\bm{\mu}_{q,N})$ is the forward kinematics mapping from the mean final joint configuration to end-effector position. This constraint ensures the mean trajectory reaches the target position exactly in task space.

\subsection{Temporal constraint}
The formulated optimal control problem can be solved with or without constraining movement duration. Without a constraint, movement duration emerges from the optimization~\cite{berret2021stochastic}, but then the width of the target and subsequently the difficulty of reaching it are not accounted for. One of the contributions of our work is to impose a duration constraint to incorporate the required target width, which has been missing from the formulation until now. By formulating the temporal constraint based on Fitts' law, the planner can adapt movement duration to task difficulty.

Fitts' law relates movement time to task difficulty through the Index of Difficulty (ID), defined as
\begin{equation}
    \mathrm{ID}^{\mathrm{pred}} = \log_2\Big(\tfrac{D^{\mathrm{pred}}}{W}+1\Big),
    \label{eq:indexofdifficulty}
\end{equation}
where $D^{\mathrm{pred}}$ is the predicted distance to the target and $W$ is the assumed-known target width, which together quantify the predicted task difficulty in bits. Movement time is then predicted as
\begin{equation}
    \mathrm{MT}^{\mathrm{pred}} = a + b \cdot \mathrm{ID}^{\mathrm{pred}},
    \label{eq:fittslaw}
\end{equation}
where the parameter $a$ is the function intercept representing baseline components such as reaction time and initial movement time, which are relatively stable across different conditions. The parameter $b$ is the function slope representing the capabilities/limitations of human motor control in the face of signal-dependent noise.

The predicted movement duration is converted to a discrete horizon constraint by dividing by the time step $h$
\begin{equation}
    N = \Big\lceil \tfrac{\mathrm{MT}^{\mathrm{pred}}}{h} \Big\rceil,
    \label{eq:horizonconstraint}
\end{equation}
where $\left\lceil \cdot \right\rceil$ denotes the ceiling function, which rounds up to the nearest integer. This horizon constraint determines the number of optimization steps, effectively setting the temporal structure of the planned trajectory based on task difficulty. By integrating Fitts' law into the planning framework, the model captures how humans naturally adjust movement speed according to the accuracy requirements and distance of the reaching task.

\subsection{Iterative Fitts' law parameter adaptation}
The horizon constraint presented predicts movement duration using fixed Fitts' law parameters $a$ and $b$.
Typically, $a$ and $b$ are fixed values based on the capabilities/limitations of human motor control. However, during co-manipulation with a robot, the robot can artificially augment the capabilities beyond normal human limitations. Thus, fixed parameters $a$ and $b$ cannot account for the expected speed increase resulting from robot augmentation. The robot's precision in the main movement direction effectively loosens accuracy requirements in the authority handover plane dimensions, which is expected to enable faster human movements according to the speed-accuracy trade-off. Consequently, the throughput of the combined human-robot system, defined as the rate at which the system can complete movements of a given difficulty, is expected to increase compared with human-only performance.

To capture this predicted increase in throughput, we created a mechanism to adapt the original Fitts' law parameters based on observed behavior during human-robot collaboration, with $b$ being the critical parameter. The parameters are initialized with baseline values representing typical human-only performance. Then, a Recursive Least Squares (RLS) algorithm is used to update the parameters after each movement. Unlike the prediction phase, which uses predicted values (\ref{eq:indexofdifficulty}), the parameter adaptation uses actual observed measurements from completed movements. The algorithm is based on the linear regression model defined as
\begin{equation}
    \mathrm{MT}^{\mathrm{act}}_i
    = a + b \log_2\!\Big(\tfrac{D^{\mathrm{act}}_i}{W} + 1\Big) + \varepsilon_i
    = \boldsymbol{\phi}_i^\top \theta + \varepsilon_i,
    \label{eq:rls_model}
\end{equation}
where $\mathrm{MT}^{\mathrm{act}}_i$ is the actual movement duration for the i-th movement, $D^{\mathrm{act}}_i$ is the actual Euclidean distance moved and $W$ is the assumed-known target width. The term $\varepsilon_i$ represents the deviation of the observed movement time from the linear prediction, including both random variability and systematic error from the parameter vector $\theta$ not having converged to its true value.

By iteratively minimizing this error through parameter updates, the RLS algorithm drives $\theta$ towards its true value. The regressor vector $\boldsymbol{\phi}_i$ and parameter vector $\theta$ are defined as
\begin{equation}
  \boldsymbol{\phi}_i =
  \begin{bmatrix}
    1 \\[3pt]
    \log_2\!\big(\tfrac{D^{\mathrm{act}}_i}{W} + 1\big)
  \end{bmatrix},
  \qquad
  \theta =
  \begin{bmatrix}
    a \\[3pt] b
  \end{bmatrix},
\end{equation}
where the first element of $\boldsymbol{\phi}_i$ corresponds to the intercept $a$, and the second element corresponds to slope $b$. After each completed movement $i$, the RLS algorithm performs three sequential updates. First, the gain vector is computed as
\begin{subequations}
    \label{eq:rls_updates}
\begin{equation}
    K_i = \frac{P_{i-1}\,\boldsymbol{\phi}_i}{\lambda + \boldsymbol{\phi}_i^\top P_{i-1} \boldsymbol{\phi}_i},
    \label{eq:rls_gain}
\end{equation}
which determines how strongly the new observation should influence the parameter update based on the previous uncertainty estimate $P_{i-1}$ and the current regressor $\boldsymbol{\phi}_i$. Second, the parameter estimate is updated as
\begin{equation}
    \theta_i = \theta_{i-1} + K_i \big(\mathrm{MT}^{\mathrm{act}}_i - \boldsymbol{\phi}_i^\top \theta_{i-1}\big),
    \label{eq:rls_update}
\end{equation}
where the prediction error $\big(\mathrm{MT}^{\mathrm{act}}_i - \boldsymbol{\phi}_i^\top \theta_{i-1}\big)$  represents the difference between the observed movement time and the time predicted by the current parameters. This error, weighted by the gain $K_i$, adjusts the parameters toward values that better explain the observed behavior. Third, the covariance matrix is updated as
\begin{equation}
      P_i = \tfrac{1}{\lambda}\Big(P_{i-1} - K_i \boldsymbol{\phi}_i^\top P_{i-1}\Big),
    \label{eq:rls_covariance}
\end{equation}
\end{subequations}
which tracks the estimation uncertainty for use in the next iteration. The forgetting factor $\lambda \in (0,1]$ controls adaptation speed. Values of $\lambda$ closer to 1 result in slower adaptation that weighs all past observations more equally, while smaller values enable faster adaptation by discounting older observations. Through this iterative process, the planner gradually refines the Fitts' law parameters to reflect how the human-robot system completes movements of a given difficulty, enabling the adaptation to the predicted faster movements of humans in response to the robot's accuracy augmentation.

\section{Experiments}
\label{sec:experiments}

\subsection{Simulation: Setup} 
The goal of the simulations was to validate research objectives O1 and O2. Simulations were performed using the KUKA LBR iiwa 7 R800 robot manipulator, which was also used in the physical experiment. Two sets of reaching movements were simulated to evaluate how movement duration adapts to task difficulty. We generated movements from a fixed initial position similar to that shown in Fig.~\ref{fig:overview}, moving in the downward direction. Set A tested movements with four different distances (0.15, 0.30, 0.45, 0.60) m to a constant target width of 0.10 m. Set B tested movements with a constant distance of 0.311 m to four different target widths (0.025, 0.05, 0.1, 0.2) m. For each movement, we recorded end-effector trajectories, duration, and computation time.

Trajectories were generated from a fixed home configuration $\bm{\mu}_{0}$ with zero initial velocities. For each simulation, the authority handover plane was positioned with respect to the initial position to achieve the required test distance. Following the approach described in the final state constraint formulation, the target position $\bm{p}_{\mathrm{target}}$ was determined as the closest position on this plane to the end-effector. The final state constraint enforced the end-effector at the target position $\bm{p}_{\mathrm{target}}$ as formulated in the framework, while additionally requiring zero final velocities. The horizon constraint (\ref{eq:horizonconstraint}) was applied throughout with fixed Fitts' law $a=0.65$ and $b=0.12$. The RLS adaptation scheme was not applied, as no actual movement durations or distances were available to drive parameter updates. The resulting nonlinear program was solved using IPOPT with the MUMPS linear solver.

\subsection{Simulation: Results}

\begin{figure*}
    \centering
    \includegraphics[width=0.49\textwidth]{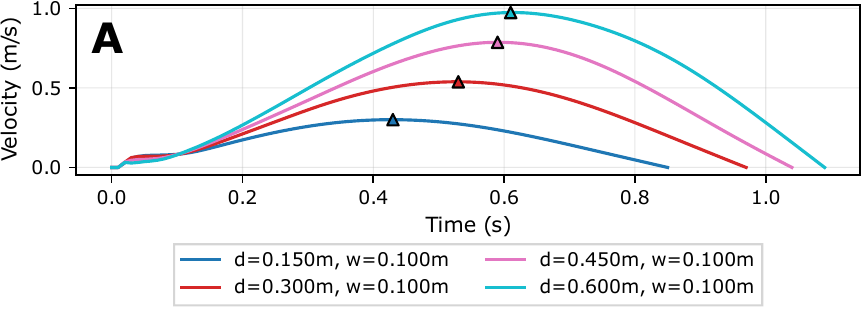}
    \hfill
    \includegraphics[width=0.49\textwidth]{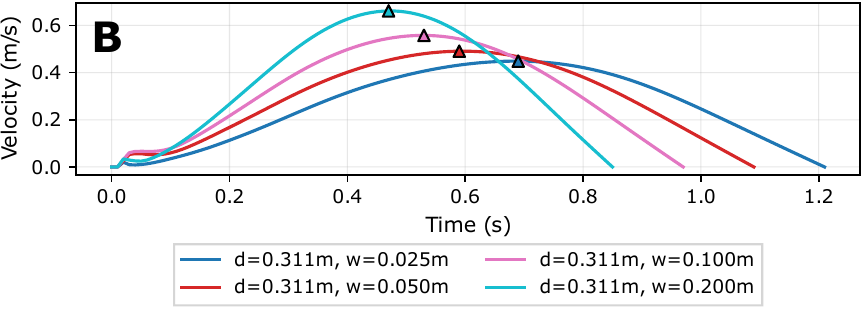}
    \caption{End-effector velocity profiles for both sets of simulated tasks: Set A (left) and Set B (right).}
    \label{fig:velocity_profiles}
\end{figure*}

Figure~\ref{fig:velocity_profiles} shows endpoint velocity profiles for both tested sets. All profiles exhibit bell-shaped curves with single peaks. The optimizer converged for all movements with the end-effector reaching the target center and zero final velocities. Table~\ref{tab:task_stats} presents movement durations and solve times. In Set A, movement durations ranged from 0.85 s to 1.09 s, with solve times 2.41$\pm$0.63 s. In Set B, duration ranged from 0.85 s to 1.21 s, with solve times averaging 2.23$\pm$0.40 s.
\begin{table}[h]
  \centering
  \footnotesize
  \caption{Task statistics for movements with varying distance and width.}
  \label{tab:task_stats}
  \begin{tabular}{cccc}
    \toprule
    \multicolumn{4}{c}{\textbf{Set A: fixed width W=0.100 m, varying distance}} \\
        \midrule
        Distance (m) & Width (m) & Duration (s) & Solve time (s) \\
        0.150 & 0.100 & 0.8500 & 1.6802 \\
        0.300 & 0.100 & 0.9700 & 2.0249 \\
        0.450 & 0.100 & 1.0400 & 2.6004 \\
        0.600 & 0.100 & 1.0900 & 3.3500 \\
        --    & --    & $0.9875 \pm 0.0901$ & $2.4139 \pm 0.6326$ \\
        \midrule 
    \multicolumn{4}{c}{\textbf{Set B: fixed distance D=0.311 m, varying width}} \\
    \midrule
    Distance (m) & Width (m) & Duration (s) & Solve time (s) \\
    0.311 & 0.025 & 1.2100 & 2.7808 \\
    0.311 & 0.050 & 1.0900 & 2.4401 \\
    0.311 & 0.100 & 0.9700 & 1.8550 \\
    0.311 & 0.200 & 0.8500 & 1.8366 \\
    --    & --    & $1.0300 \pm 0.1342$ & $2.2281 \pm 0.4009$ \\
    \bottomrule
  \end{tabular}
\end{table}

\subsection{Experiment: Setup}
To validate research objective O3, we conducted a physical experiment with a human participant performing reaching movements with a KUKA LBR iiwa 7 R800 robotic arm. The task was the same as described in the introduction and illustrated in Fig.~\ref{fig:overview}. Two experimental conditions were tested with 15 trials each. With the baseline condition A, we tested human reaching movements without robot assistance, where the robot stiffness was zero to allow free movements, but still measured them. With Condition B, we tested human reaching movements, where the robot augmented accuracy and speed using the proposed method.

The task required moving from a fixed initial position to a target position on the authority handover plane as quickly as possible while strictly avoiding overshoot. The participant was explicitly instructed that overshooting beyond the plane would represent a collision with the working plane below. For each trial, a target position was randomly generated on the authority handover plane and shown on a monitor in front of the participant. In a real collaborative task, the human would determine the target position based on task knowledge without the system knowing it. However, for the sake of systematic evaluation, here the target position was generated externally by the system so that we could quantitatively know where the target was. This randomly generated target position $\bm{p}_{true}$ represents the actual handover position that the human had to aim for. Note that this differs from $\bm{p}_{target}$ used in the planning framework. The participant was free to position the end-effector at any initial location before each trial, resulting in varying distances to the handover plane and Indices of Difficulty across trials.

In each trial, we recorded movement duration and Euclidean distance traveled. We determined the movement duration by timing the start of the movement when the human displaced the end-effector from the initial position and the end of the movement when the end-effector first came within the target width $W$ of $\bm{p}_{\mathrm{true}}$. The movement was considered complete only if the end-effector remained within the target region for a threshold duration. 

We also recorded the evolution of Fitts' law parameters $a$ and $b$ across trials to evaluate the adaptation process of the RLS algorithm. The planner configuration and parameters were identical to those used in the simulation experiment, except for the Fitts' law parameters, which were adapted. Both conditions used the same initial Fitts' law parameters ($a=0.3, b=0.35$) determined from preliminary tests without the planner enabled. Unlike the simulation experiment where fixed parameters were used, here the RLS adaptation mechanism was applied after each movement to update the Fitts' law parameters online based on observed movement durations and distances. The assumed-known target width was set to 0.1 m and remained constant across all trials.

In Condition A, initially, the participant freely guided the end-effector to the desired initial position and then pressed key 'P' to switch the impedance controller to high stiffness (500 N/m in all directions). The robot then maintained that position to establish a clear starting point for timing the movement duration. The participant then initiated movement by displacing the end-effector from this high-stiffness state, which automatically started the timer and switched the controller back to low-stiffness mode. The online adapted parameters of Fitts' law were recorded for comparison purposes but not used for trajectory planning.

In Condition B, similarly to above, the participant first moved the robot into the desired initial position. This also triggered trajectory planning toward the target position $\bm{p}_{\mathrm{target}}$. Once the trajectory was planned and shown on the monitor, the participant pressed 'C' to confirm and load the trajectory into the impedance controller. The participant then initiated movement by displacing it from the initial position, which automatically started the timer and switched the impedance to execution mode. During trajectory execution mode, the controller applied high stiffness (800 N/m) orthogonal to the authority handover plane in the main movement direction (z-axis) to augment human accuracy, while maintaining low stiffness (0.1 N/m) within the plane (x-y axes) to permit the human to control the exact authority handover position on the handover plane (Fig. \ref{fig:overview}). The robot executed the planned trajectory in the main movement direction, while the participant controlled the in-plane position.

\begin{figure}[t]
    \centering
    \includegraphics[width=0.9\linewidth]{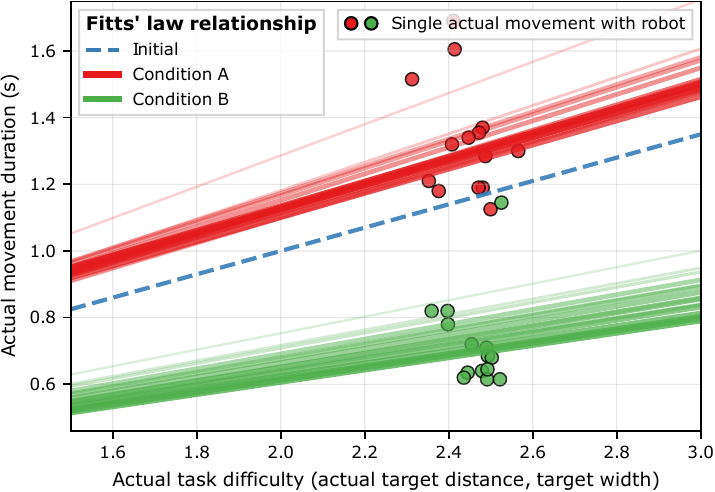}
    \caption{Measured movement data and RLS parameter adaptation for baseline (Condition A, red) and the proposed method (Condition B, green). Each dot represents a single trial, and each line shows the Fitts' law prediction after an RLS update. The initial Fitts' law relationship is indicated by the blue dotted line (a = 0.3, b = 0.35).}
    \label{fig:throughput_capacity}
    %\vspace{-4mm}
\end{figure}

By moving faster or slower than predicted and working against the stiff impedance controller, the participant could finish movements earlier or later than the planned duration. This allowed the RLS algorithm to adapt the parameters based on actual performance. Once the movement was confirmed as complete, impedance switched to low stiffness, and the participant was free to reposition to the desired initial position for the next trial. In Condition B, the online adapted parameters of Fitts' law were then used to plan the trajectory for the subsequent trial.

\subsection{Experiment: Results}
The participant performed 15 trials for each condition. The recorded trials captured a range of task difficulties from 2.31 to 2.57 bits, corresponding to variation in distances between 0.42 and 0.56 m and a constant target width of 0.1 m. Movement data and RLS parameter updates for both conditions are shown in Fig. \ref{fig:throughput_capacity}. The results show that the proposed augmentation method (Condition B, green) decreased the movement duration across all task difficulties, compared to the case without it (Condition A, red). The robot with the planner framework was able to augment the human speed without sacrificing accuracy.

\begin{figure}[t]
\centering
\includegraphics[width=0.9\linewidth]{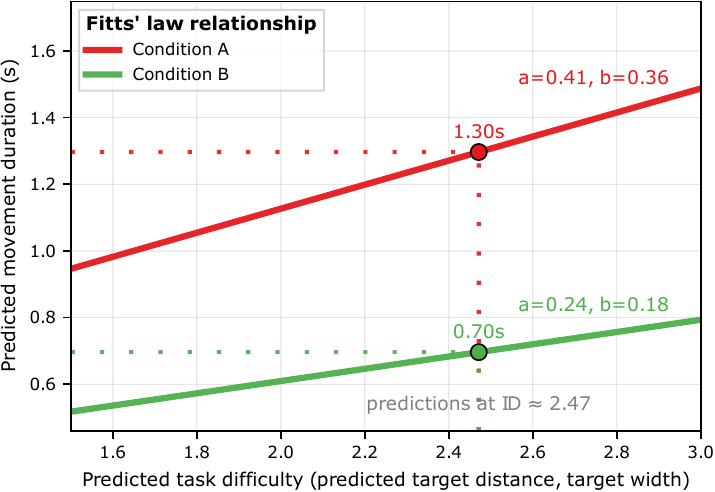}
\caption{Final Fitts' law relationships after 15 RLS updates for both conditions, showing predicted movement duration at constant task difficulty (ID = 2.47).}
\label{fig:predicted_movement_duration}
\end{figure}

For a closer analysis, the final converged parameter values of Fitts' law functions after 15 RLS updates for both conditions are shown in Fig.~\ref{fig:predicted_movement_duration}. The predicted movement duration is shown for a constant task difficulty (ID = 2.47) to demonstrate the difference between conditions after convergence. Condition A converged to parameter values $a = 0.41$ and $b = 0.36$, predicting a movement duration of 1.30 s. Condition B converged to parameter values $a = 0.24$ and $b = 0.18$, predicting a movement duration of 0.70 s. There was a 46\% reduction in predicted movement duration when using the proposed method. The throughput parameter $b$ decreased by 50\%, while parameter $a$ decreased by 41\%.

\section{Discussion}
This work developed the first stochastic sensorimotor planning framework for human-robot co-manipulation that integrates accuracy and speed augmentation. The framework suppresses endpoint variability through spatial constraints to provide accuracy augmentation, while a Fitts' law-based temporal constraint adapts movement duration to anticipate and synchronize with speed increases resulting from this augmentation.

The simulation experiment successfully demonstrated that Objective O1 and Objective O2 were achieved. The planner generated smooth trajectories with bell-shaped velocity profiles that conform with the neuroscience literature~\cite{flash1985coordination,franklin2011computational,peternel2017unifying}. Movement duration was adapted appropriately to task difficulty by increasing with distance and decreasing with target width, which is consistent with Fitts' law~\cite{fitts1954information}. This indicates that our method successfully incorporated SAT into the planning. Finally, thanks to the final state constraint, movement terminated exactly at the target center, which can help with the desired augmentation in a human-robot co-manipulation task. 

The human-robot co-manipulation experiment demonstrated that Objective O3 was successfully achieved. The human participant was able to go beyond the typical limitations of human motor control and moved faster with the robot augmentation, as seen in Figure \ref{fig:throughput_capacity}. The combined human-robot system adapted to these faster movements through the horizon constraint. Parameters $a$ and $b$ decreased consistently through the trials, thanks to the iterative Fitts’ law parameter adaptation mechanism. The 50\% reduction in $b$ reflects the increased system throughput, as lower $b$ values correspond to higher information processing rates measured in bits per second. This was enabled by accuracy augmentation and resulted in performance enhancement, where the human-robot system could complete movements of a given Index of Difficulty more rapidly.

The unexpected 41\% decrease in parameter $a$ (from 0.41 to 0.24) appears inconsistent with theory. Parameter $a$ represents the inherent human reaction and movement initiation time, which should remain relatively constant across contexts. One possible explanation for this is that the range of tested task difficulties (2.31-2.57 bits) was too small for the RLS algorithm to independently estimate the intercept and slope based on the increased movement speed. Nevertheless, the resulting 46\% reduction in predicted movement duration (from 1.30 s to 0.70 s at ID = 2.47) demonstrates that the planner successfully adapted to the faster movements.

Previous theory-driven sensorimotor planning approaches for co-manipulation focused on conforming to standard human motor behavior and limitations~\cite{haninger2024planning}. On the other hand, this study shows that robots can augment humans beyond the standard behavior and limitations by anticipating and adapting to the changed behavior that emerges from robot assistance. When there are high consequences of missing the target in the main movement direction, e.g., hitting a surface of a delicate object located in a plane, the human alone would need to be slow to be precise and avoid overshoots in the axis normal to the plane. Incorporating human motor control modeling into robot planning enables insights about the target that can be used for the robot to make sure the movement is precise in that axis. Humans can then speed up knowing that the robot will not allow dangerous overshoots, especially when the consequences of being imprecise in the remaining directions are lower, e.g., in-plane axes, where fine adjustment can be made at the end without danger. The observed speed increase in these proof-of-concept experiments provides evidence for this hypothesis.

This work focused on predicting movement duration based on task difficulty and the increased capability of the human-robot system. What was not yet addressed is how to also incorporate context-dependent factors, such as the subjective value of time and the consequences of missing. For example, meeting production demands might influence the importance of execution time, and augmentation can be strengthened to reduce time. On the other hand, the consequences of missing the target are not always the same. If we are dealing with fragile or sharp objects, there is more contextual risk, and the human might need to be more careful and thus slower. In future, we can explicitly incorporate these external factors into the Fitts' law adaptation.

These proof-of-concept experiments validated the feasibility of the proposed method and highlighted its potential for augmentation in human-robot co-manipulation. The main limitation of the current study is that the narrow range of task difficulties tested limits the robustness of the parameter estimates. Future work should focus on increasing the tested range of distances and target widths to provide more robust parameter estimates for a specific user. Furthermore, the same should also be tested across multiple participants to gain insights into variability between different users.

\section*{Appendix}
\label{FirstAppendix}
\begin{figure*}[t]
    \centering
    \includegraphics[width=0.95\linewidth]{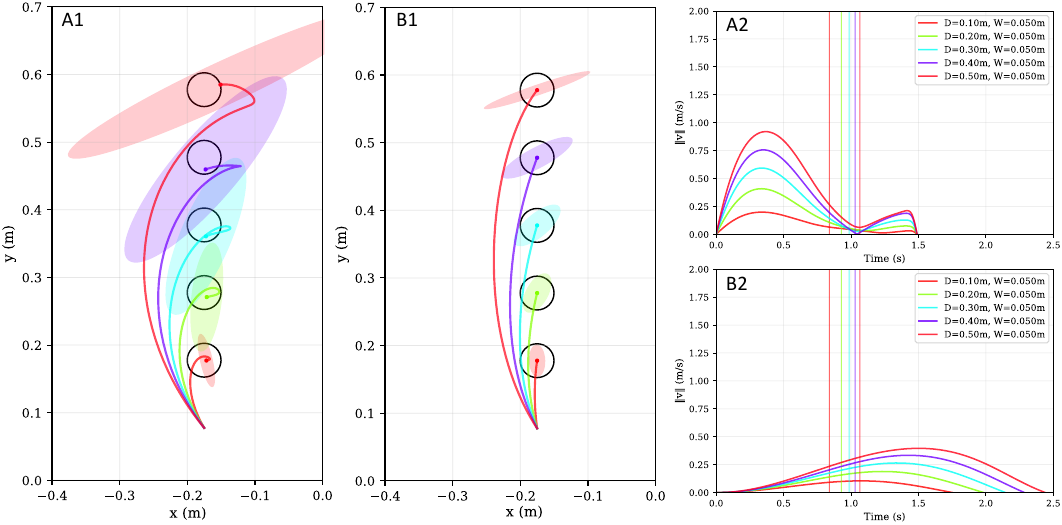}
    \caption{Endpoint trajectories (left) and velocity profiles (right) from Model A~\cite{peternel2017unifying} and Model B~\cite{berret2021stochastic} for targets (black circles) at five different distances and a constant target width. The 2D ellipses represent the predicted endpoint covariance from signal-dependent noise, not distributions from multiple trajectory samples. Vertical lines in the velocity profiles indicate durations predicted by Fitts’ law with parameters $a=0.65$, $b=0.12$.}
    \label{fig:comparison}
    %\vspace{-4mm}
\end{figure*}

Figure \ref{fig:comparison} shows comparison performed in simulation between Model A~\cite{peternel2017unifying} (code adopted from~\cite{haninger2024planning}) and Model B~\cite{berret2021stochastic}. Model A exhibits endpoint variability at the target as a result of optimizing perceived benefit by reducing time spent on movements at the expense of some probability of missing. It is interesting to observe clear final corrective actions, which were not observed in previous studies of this specific model. Also, smaller distances (low difficulty) do not require the extra final corrective actions, and the initial ballistic sub-movements are already sufficient to reach the target, which is in line with observed empirical human data~\cite{peternel2019target}. This would imply that Model A inherently mimics a multi-component reaching strategy~\cite{elliott2001century,peternel2019target}, where the initial ballistic component is fast and imprecise to get into the vicinity of the target as quickly as possible, while the final corrective component is slow and precise to ensure the target is reached. On the other hand, Model B clearly exhibits the final state constraints and goes into the target directly. With robot accuracy augmentation, information about the probability of missing and a multi-component reaching strategy to deal with it may not play a key role, and enforcing the final state constraint is more beneficial. For that reason, Model B was considered more suitable for robots augmenting humans in co-manipulation.

%\section*{Acknowledgement}
%{\small }

\bibliographystyle{IEEEtran}
\bibliography{arXiv2026}

\end{document}